\documentclass{article}

\usepackage{microtype}
\usepackage{graphicx}
\usepackage{subfigure}
\usepackage{booktabs} % for professional tables

\usepackage{hyperref}
\usepackage{bbm}

\newcommand{\norm}[1]{\left \lVert #1 \right \rVert}

\usepackage[accepted]{icml2025}

\usepackage{amsmath}
\usepackage{amssymb}
\usepackage{mathtools}
\usepackage{amsthm}

\usepackage[capitalize,noabbrev]{cleveref}

\theoremstyle{plain}
\newtheorem{theorem}{Theorem}[section]

\theoremstyle{definition}
\newtheorem{definition}[theorem]{Definition}

\theoremstyle{remark}

\usepackage[textsize=tiny]{todonotes}

\icmltitlerunning{Submission and Formatting Instructions for ICML 2025}

\begin{document}

\twocolumn[
\icmltitle{Robustness questions the interpretability of graph neural networks: what to do?}



\icmlsetsymbol{equal}{*}

\begin{icmlauthorlist}
% \icmlauthor{Anonymous}{}%{equal,yyy}
\icmlauthor{Kirill Lukyanov}{tai,isp,mipt}
\icmlauthor{Georgii Sazonov}{isp,msu}
\icmlauthor{Serafim Boyarsky}{sad}
\icmlauthor{Ilya Makarov}{tai,airi}
\end{icmlauthorlist}

\icmlaffiliation{tai}{ISP RAS Research Center for Trusted Artificial Intelligence, 109004 Moscow, Russia}
\icmlaffiliation{isp}{Ivannikov Institute for System Programming of the Russian Academy of Sciences, 109004 Moscow, Russia}
\icmlaffiliation{mipt}{Moscow Institute of Physics and Technology (National Research University), 141700 Moscow, Russia}
\icmlaffiliation{msu}{Lomonosov Moscow State University, Leninskie Gory, 1, Moscow, 119991, Russia}
\icmlaffiliation{airi}{AIRI, 121170 Moscow, Russia}
\icmlaffiliation{sad}{Yandex School of Data Analysis,
bld. 2, 11, Timur Frunze st., Moscow, 119021, Russia}

% \icmlcorrespondingauthor{Anonymous}{}%{first1.last1@xxx.edu}
\icmlcorrespondingauthor{Kirill Lukyanov}{lukyanov.k@ispras.ru}

\icmlkeywords{Machine Learning}

\vskip 0.3in
]



\printAffiliationsAndNotice{}  % leave blank if no need to mention the equal contribution

\begin{abstract}
    Graph Neural Networks (GNNs) have become a cornerstone in graph-based data analysis, with applications in diverse domains such as bioinformatics, social networks, and recommendation systems. However, the interplay between model interpretability and robustness remains poorly understood, especially under adversarial scenarios like poisoning and evasion attacks. This paper presents a comprehensive benchmark to systematically analyze the impact of various factors on the interpretability of GNNs, including the influence of robustness-enhancing defense mechanisms.

    We evaluate six GNN architectures based on GCN, SAGE, GIN, and GAT across five datasets from two distinct domains, employing four interpretability metrics: Fidelity, Stability, Consistency, and Sparsity. Our study examines how defenses against poisoning and evasion attacks, applied before and during model training, affect interpretability and highlights critical trade-offs between robustness and interpretability. The framework will be published as open source.

    The results reveal significant variations in interpretability depending on the chosen defense methods and model architecture characteristics. By establishing a standardized benchmark, this work provides a foundation for developing GNNs that are both robust to adversarial threats and interpretable, facilitating trust in their deployment in sensitive applications.
\end{abstract}

\section{Introduction}
% Graph Neural Networks (GNNs) have been attracting significant attention recently, as they have demonstrated their applicability across various domains where data possesses an inherent graph structure. Proper processing of this structure allows for more accurate predictions. However, GNN deployment in high-stakes fields, as for other families of machine learning algorithms, necessitates meeting various conditions of trustworthiness (such as robustness, explainability, privacy, etc.),  determined by the specific application setting. Consequently, there is a growing body of research on GNNs focusing on their explainability, as well as on methods for constructing GNNs that are robust to adversarial attacks. However, the simultaneous fulfillment of explainability and robustness requirements remains an underexplored area. This is a critical issue, as efforts to defend models against such attacks may compromise the correctness or reduce the quality of their interpretations, creating challenges in achieving both interpretability and security. This study focuses on analyzing this relationship, investigating how the integration of defense mechanisms affects the quality of interpretations.

Graph Neural Networks (GNNs) have rapidly emerged as a powerful tool for analyzing graph-structured data, driving progress in fields such as bioinformatics, social networks, and recommendation systems. Their ability to capture complex relational structures has positioned GNNs at the forefront of machine learning research. However, as these models are increasingly adopted in critical domains, the need for trustworthy predictions—encompassing both interpretability and robustness to adversarial threats—has become more pressing.

\begin{figure*}[ht]
    \vskip 0.2in
    \begin{center}
        \centerline{\includegraphics[width=1.8\columnwidth]{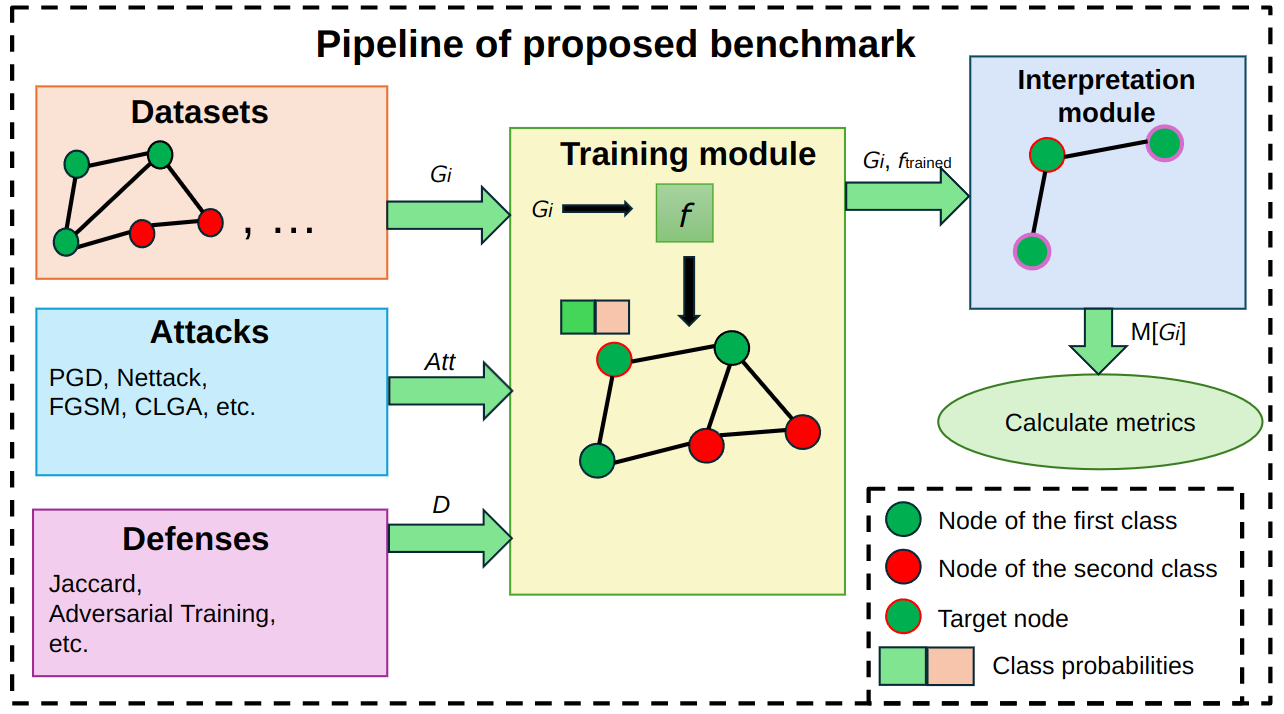}}
        \caption{Overall Benchmark Pipeline. One of the available datasets \( G_i \) can be selected, along with an attack method \( \text{Att} \) and a defense method \( D \). The next stage involves training a GNN \( f \) on the selected graph \( G_i \) while applying the chosen attack and defense methods. The trained model \( f^{\text{trained}} \) and the dataset \( G_i \) are then passed to the interpretation module, where one of the available interpretation methods can be applied to generate a mask \( M(G_i) \) that highlights the important subgraph. Based on the generated masks, interpretability metrics can be computed. A detailed description of all definitions is provided in Section \ref{section_methods}.}
        \label{img:pipeline}
    \end{center}
    \vskip -0.2in
\end{figure*}

Despite significant advances in interpretability and robustness, ensuring both properties simultaneously in GNNs remains a challenge. Numerous defense mechanisms have been introduced to counteract adversarial attacks~\cite{goodfellow2014explaining, finlay2021scaleable, guo2017countering, wu2019adversarial, zhang2020gnnguard}, while various interpretability techniques have been developed~\cite{ying2019gnnexplainer, funke2020hard, yuan2021explainability, zhang2022protgnn, Dai2022ACS}. However, two key limitations hinder their widespread applicability. Computational Complexity: Graph-based interpretability methods often exhibit prohibitively high computational costs. For instance, widely used techniques such as Zorro~\cite{funke2020hard} and SubgraphX~\cite{yuan2021explainability} can require 1-3 days to interpret a single node in datasets like Photo and Computers from the Amazon dataset family~\cite{mcauley2015image}. Architectural Constraints: Many existing methods impose specific architectural requirements, reducing their generalizability. For example, RobustGCN~\cite{zhu2019robust} relies on specialized convolutional layers that support differentiation concerning the adjacency matrix, diverging from the gradient propagation approach in PyTorch-Geometric~\cite{Fey2019FastGR}. ProtGNN~\cite{zhang2022protgnn} requires adding a prototype layer at the end of the model, and its available implementation \url{https://github.com/zaixizhang/ProtGNN} supports only graph classification tasks. These constraints limit the applicability and evaluation of many methods, making it difficult to develop generalizable solutions for trustworthy GNNs. 

A more fundamental issue is the lack of research on the relationship between robustness and interpretability. This challenge extends beyond graph-based models and is prevalent in the broader field of trustworthy AI, where most studies treat these properties as isolated objectives. Few works have explored the interplay between different trustworthiness criteria. One study~\cite{Moshkovitz2021ConnectingIA} analyzed interpretability and robustness but focused on simple models like decision trees and small tabular datasets. Another work~\cite{Szyller2022ConflictingIA} examined conflicts arising when evasion and poisoning attacks occur simultaneously, but only in the image domain. We argue that trustworthiness should be approached as a multi-objective problem, balancing multiple criteria rather than optimizing for a single property in isolation.

In this work, we systematically examine how various factors influence the interpretability of GNNs, using four key metrics: Fidelity, Stability, Consistency, and Sparsity. Specifically, we address the following research questions:
\begin{enumerate}
    \item How do structural and domain-specific properties of graphs influence interpretability?
    \item How do GNN architectural choices, such as the number and type of convolutional layers, affect interpretability quality?
    \item How do defense mechanisms against poisoning and evasion attacks impact interpretability?
\end{enumerate}

By framing trustworthiness as a multi-objective challenge, we aim to provide insights that can guide the development of more generalizable and computationally feasible approaches for interpretable and robust GNNs.

The contributions of this work are as follows:
\begin{itemize}
    \item We showed that Consistency and Fidelity are more suitable for evaluating interpretation methods, as they remain stable. In contrast, Sparsity and Stability are more sensitive to modifications, making them better suited for assessing how even small changes affect interpretability. 
    \item We showed that most defense mechanisms improve interpretability, but some have a less positive impact compared to others.
    \item We highlighted the limitations of existing interpretability metrics, emphasizing the need for refinement to better capture the effects of model modifications.
\end{itemize}

The rest of the paper is organized as follows. Section \ref{section_review} reviews the relevant methods and explains the rationale behind their selection. Section \ref{section_methods} presents the formal problem statement and evaluation approach. Section \ref{section_experiments} outlines the experimental methodology, with results discussed in Section \ref{section_discussion}. Finally, Section \ref{section_conclusion} summarizes the findings and discusses potential directions for future work.

\section{Review}
\label{section_review}

This section provides an overview of existing approaches to interpretability and robustness, explaining the rationale behind selecting specific directions and methods as the most suitable for addressing the research questions.

\subsection{Approaches to machine learning models attacks and defense}

Machine learning models are vulnerable to various attacks, each targeting different aspects of their functionality. These include privacy attacks~\cite{Olatunji2021MembershipIA, shaikhelislamov2024study}, adversarial attacks~\cite{Zgner2018AdversarialAO}, and others~\cite{Pal2020ActiveThiefME}, each posing unique challenges and requiring specialized defenses. Privacy attacks aim to extract sensitive information, while model extraction attacks attempt to replicate the model’s functionality. Adversarial attacks exploit model vulnerabilities to manipulate predictions. This work focuses on defense mechanisms against adversarial attacks.

Adversarial attacks are classified into poisoning attacks~\cite{Zhang2022UnsupervisedGP}, evasion attacks~\cite{Zgner2018AdversarialAO}, backdoor attacks~\cite{Zheng2022MotifBackdoorRT}, and others~\cite{dombrowski2019explanations}. Poisoning attacks compromise the model during training by injecting malicious data, while evasion attacks perturb inputs at inference time to induce incorrect predictions. Backdoor attacks embed hidden triggers to alter behavior under specific conditions. This work examines defense strategies against poisoning and evasion attacks, as they are the most commonly studied.

Attacks on GNNs present unique challenges compared to traditional models. In addition to manipulating feature matrices~\cite{madry2017towards}, adversaries can target the graph structure by adding, removing, or modifying edges and nodes~\cite{Zhang2022UnsupervisedGP}. This dual attack surface complicates the development of robust defenses and requires specialized methods tailored to the graph domain.

To evaluate how defense mechanisms impact the interpretability of GNNs, we selected either state-of-the-art (SOTA) or widely adopted defense methods. This selection ensures that the analysis reflects the latest advancements and commonly used practices in defending against adversarial threats.

\subsection{Approaches to machine learning models interpretability}

Machine learning model interpretability can be categorized into: post-hoc interpretability, self-interpreting models, and counterfactual explanations~\cite{Dai2022ACS}. Post-hoc methods explain the behavior of pre-trained models without modifying their architecture, making them broadly applicable. Self-interpreting models prioritize transparency through specific architectural choices~\cite{zhang2022protgnn, Han2022GlobalCI}, but lack flexibility for arbitrary architectures. Counterfactual explanations identify minimal input changes that alter predictions, providing insights into decision boundaries~\cite{Lucic2021CFGNNExplainerCE, Verma2020CounterfactualEA}. However, counterfactual methods are less suitable for benchmarking due to difficulties in defining metrics for systematic evaluation. This work focuses on post-hoc methods, which enable the analysis of pre-existing models without architectural constraints and support robustness evaluation.

GNNExplainer~\cite{ying2019gnnexplainer} was selected as the primary interpretability method. Despite being introduced some time ago, it remains one of the most effective approaches due to its relatively fast execution and lack of dependency on specific architectural choices. Additionally, SubgraphX~\cite{yuan2021explainability} was considered as a supplementary interpretability method.

\subsection{Evaluation of interpretability}

Metrics are crucial for building benchmarks and systematically comparing factors influencing interpretability outcomes~\cite{DoshiVelez2017TowardsAR}. In this study, we selected four key metrics: Fidelity, Stability, Consistency, and Sparsity.

While other interpretability metrics have been proposed, they are not considered here. The selected metrics were chosen based on recommendations in the literature. According to~\cite{DoshiVelez2017TowardsAR}, objective metrics help avoid reliance on subjective evaluations and improve result reproducibility. These metrics are widely used and can be automatically calculated from the model’s outputs, simplifying the process and reducing computational complexity. Expert evaluations are often inconsistent and depend on experience, making automated metrics preferable~\cite{Lipton2018}. These metrics do not require interpretable models or auxiliary classifiers, as highlighted in~\cite{Ribeiro2016}.

The selected metrics cover different aspects of interpretability: Fidelity measures how accurately the explanation reflects the model's decisions; Stability and Consistency assess the robustness and agreement of explanations; and Sparsity reduces cognitive load by minimizing the number of features in the explanation. This approach aligns with recommendations for multidimensional interpretability analysis~\cite{Guidotti2018, Miller2019}.

\subsection{Interactions among defense and interpretation methods in pipeline}

This subsection further examines problem formulations at the intersection of interpretability and robustness, providing a detailed breakdown of the pipeline.

Recent studies have explored how interpretability can aid in identifying new attack strategies and designing effective defense mechanisms. For instance,~\cite{liu2022interpretability} demonstrated that interpretability methods could reveal vulnerabilities in GNNs, leading to more robust defense strategies. In this approach, the model is first trained, followed by the application of an interpretability method to identify potential weaknesses.

Research has also addressed adversarial attacks that specifically target interpretability methods. These studies highlight how explanations generated by models can be manipulated to mislead users or obscure malicious activity. A notable contribution by~\cite{dombrowski2019explanations} examined how adversarial perturbations could degrade interpretability results by producing deceptive explanations, ultimately undermining trust in post-hoc interpretability methods. In such cases, attacks are directed at the interpretability process itself and are applied after model training but before interpretation at a specific node, similar to evasion attacks.

In our study, we focus on how various factors influence post-hoc interpretability. Depending on the experiment, a defense mechanism against evasion attacks, poisoning attacks, or no defense at all is applied. The poisoning defense mechanism is implemented before model training, while the evasion defense mechanism is applied during training. Once training is complete, the interpretability method is used. A detailed pipeline is presented in Figure \ref{img:pipeline}.

Summarizing the review, we focused on post-hoc interpretability methods while examining the impact of defense mechanisms against poisoning and evasion attacks. The architectures were selected based on their popularity, solution quality, and diverse approaches to information aggregation within graph structures. Given the specifics of the chosen methods and approaches, a sequence of their application was established to address the research questions. The next section presents the formal problem definition and evaluation methodology.

\section{Methods}
\label{section_methods}
In this section, we formally discuss a problem statement and metrics

\subsection{Problem statement}

Given the input data as a graph \( G = (X, A) \), where \( X \) is the feature matrix and \( A \) is the adjacency matrix, a GNN model \( f \), a defense method \( D \) and an interpretation method \( I \) can be defined. The interpretation method \( I \) creates an interpretation mask \( M \). Let us define how the defense method works and what the mask \( M \), which is the result of the interpretation method \( I \), represents.

\begin{definition}[Defense Method for Graph Neural Networks]  
Let \( G = (V, E) \) be an input graph with a set of nodes \( V \) and edges \( E \), represented by a feature matrix \( X \in \mathbb{R}^{|V| \times d} \) and an adjacency matrix \( A \in \{0,1\}^{|V| \times |V|} \). Let \( f \) be the model that processes the input graph \( G \) to produce an output.

A defense method \( D \) may modify the input graph \( G \) or the model \( f \), aiming to improve robustness while preserving key characteristics.

Let \( D_G = (D_X, D_A) \) be the defense method applied to the graph, where \( D_X \) and \( D_A \) modify the feature matrix \( X \) and adjacency matrix \( A \), respectively. Alternatively, the defense method may also modify the model itself, denoted as \( f^{def} \). The modified input graph or model is then given by:
\[
X^{def} = D_X(X), \quad A^{def} = D_A(A), 
\]
\[
G^{def} = (X^{def}, A^{def}) = (D_X(X), D_A(A)),
\]
or
\[
f^{def} = D_f(f).
\]

This notation allows for the unified representation of the defended graph \( G^{def} \) and/or the defended model \( f^{def} \), enabling the evaluation of defense methods' impact on both the graph and the model's performance and interpretability metrics.  
\end{definition}

\begin{definition}[Interpretation Result in Graph Neural Networks]  
Let \( G = (V, E) \) be an input graph with a set of nodes \( V \) and edges \( E \), represented by a feature matrix \( X \in \mathbb{R}^{|V| \times d} \) and an adjacency matrix \( A \in \{0,1\}^{|V| \times |V|} \). Let \( M_X(X) \in [0,1]^{|V| \times d} \) be a mask that determines the importance of node features, and \( M_A(A) \in [0,1]^{|V| \times |V|} \) be a mask that reflects the importance of connections between nodes.  

We define a unified interpretation mask for the graph as \( M_G = M = (M_X, M_A) \), where \( M_G \) encodes both feature- and structure-level importance. The important subset of the input graph is then given by:  
\[
X^{int} = X \odot M_X, \quad A^{int} = A \odot M_A.
\]

For convenience, we introduce the interpretable subgraph notation:  
\[
G^{int} = (X^{int}, A^{int}) = (X \odot M_X, A \odot M_A).
\]

This formulation provides a unified way to describe and evaluate interpretability metrics.  
\end{definition}

\subsection{Metrics}

The computation of the interpretability metrics is now defined.

\textbf{Fidelity}
 measures how accurately an interpretation method's results reflect the original model's behavior.
\[Fidelity = \frac{1}{N}\sum_{i=1}^N |f(G^{int})-f(G_i)| \]

\textbf{Sparsity}
 evaluates how simple an explanation is, or, in other words, what percentage of features are excluded from the prediction's interpretation.
 \[
    Sparsity = \frac{\sum_{j=1}^{m} \mathbbm{1}\{M(G)_j \neq 0\}}{m} \ ,
\]
    where \( M(G)_j \) — represents the value of the interpretation mask indicating the contribution of the \( j \)-th feature, and \( m \) is the total number of features

\textbf{Stability}
 measures how similar explanations are for comparable input data. Changes in explanations are quantified by adding small noise to the original data and calculating deviations.
\[
    Stability = \frac{1}{n} \sum_{i=1}^{n} ||M(G_i) - M(G^{noise}_i)||_2 \ ,
\]
where \( G_i \) — is the original input, \( G^{noise}_i \) — is the input with small perturbations, \( ||\cdot||_2 \) — is the Euclidean norm.   

\textbf{Consistency}
 measures how similar explanations are for the same input across different runs of the model or interpretation methods.
\[
Consistency = \frac{1}{n} \sum_{i=1}^{n} \cos(M(G)_i, M(G)_{i+1}) \ ,
\]
where \(M(G)_i\) and \(M(G)_{i+1}\) — are explanations for the same input obtained from different runs.

All elements are formalized, enabling the comparison of results across different architectures, interpretation methods, and the addition of various defense methods.

\section{Experiments}
\label{section_experiments}

In this section, the technical description of the experiments is provided. In particular,  we describe the datasets and architectures of the models used in the experiments, defense and interpretation methods, and the methodology of the experiments.

\subsection{Setup of Experiments}

This subsection provides the information necessary for reproducing the experiments.

\subsubsection{Datasets}

The experiments utilized the following datasets: Cora, CiteSeer, and PubMed, which belong to the citation domain \cite{sen2008collective} and are included in the Torch-Geometric library under the Planetoid dataset group, as well as Computers and Photo from the purchase graph domain \cite{mcauley2015image}, which are also available in Torch-Geometric under the Amazon dataset group. 
More detailed information on dataset statistics is provided in Appendix \ref{app:datasets}.

\subsubsection{The architecture of GNNs models}

Six models were selected for the experiments. Four models consisted of two layers of GCN, SAGE, GAT, and GIN, respectively (this models are denoted as GNNConv-2l, where GNNConv is replaced with the corresponding convolution type). Additionally, two models included three layers of GCN and SAGE (these models are denoted as GNNConv-3l). The choice of convolution types was based on performance across the considered datasets and different data aggregation approaches \cite{zhou2020graph}. Specifically, GCN belongs to spectral convolutions, SAGE to basic spatial convolutions, and GAT to attentional spatial convolutions. GIN was introduced later, with its key idea being the generation of similar embeddings for isomorphic graphs. Appendix \ref{app:models} provided A detailed description of all model architectures.

\subsubsection{Training parameters} 

The Adam optimizer with default parameters and the NLLLoss function from the PyTorch library were used for training. No batch partitioning was applied. The number of training epochs was set to 200 to ensure high classification accuracy across all datasets.

\subsubsection{Defense and interpretation methods}

The experiments utilized defense methods against poisoning attacks: Jaccard \cite{wu2019adversarial} and GNNGuard \cite{zhang2020gnnguard}) and against evasion attacks: Distillation \cite{papernot2016distillation}, Autoencoder Defender \cite{meng2017magnet}, Adversarial Training \cite{goodfellow2014explaining}, Quantization Defender \cite{guo2017countering} and Gradient Regularization Defender \cite{finlay2021scaleable}. The primary interpretation method was GNNExplainer, as implemented in torch-geometric. Additionally, SubgraphX \cite{yuan2021explainability} was used as a supplementary method; however, due to its high computational cost, it was applied only to the Cora dataset, as it required more than two days for a single-node interpretation on other datasets. Other popular post-hoc GNN interpretation methods, such as GraphMask \cite{schlichtkrull2020interpreting} and Zorro \cite{funke2020hard}, were not used for the same reason. The hyperparameters for all methods are provided in Appendix \ref{app:int-def}.

\subsubsection{The information about averaging}

One iteration of the experiment using the GNNExplainer interpretation method involved running on 5 datasets and 6 architectures, where each architecture was applied with one of the 7 defense methods and one run without any defense methods. For each iteration and dataset, the dataset was split into training and test parts in an 80/20 ratio, respectively, and a random set of 10 nodes was fixed for all architectures and defense methods. On one hand, the independence of choice between iterations maintains the random factor, on the other hand, the same set across one iteration for all architectures ensures a fair comparison, which allows for reducing the number of required iterations and minimizing the final dispersion. The stability and consistency metrics require additional averaging for each node, which was further averaged across 5 runs for each node. When calculating the stability metric on perturbed graphs, changes of no more than 5\% of the features for all nodes were allowed, as well as the removal of no more than 5\% of all nodes from the graph.

\subsection{Results of experiments}

\subsubsection{Influence of domain factors on interpretability}

\begin{table*}[t!]
    \caption{Average interpretability metrics for models with two graph layers across different datasets. $\downarrow$ indicates that a lower metric value is better, while $\uparrow$ indicates that a higher value is better.}
    \label{tab:all_data_2l}
    \vskip 0.15in
    \begin{center}
        \begin{small}
        \begin{sc}
        \begin{tabular}{lccccc}
            \hline
            Dataset & Cora & CiteSeer & PubMed & Photo & Computers \\
            \hline
            Consistency ($\uparrow$) & 0.998 $\pm$ 0.003 & 0.997 $\pm$ 0.005 & 0.998 $\pm$ 0.002 & 0.998 $\pm$ 0.002 & 0.998 $\pm$ 0.002 \\
            Fidelity ($\uparrow$)    & 0.971 $\pm$ 0.039 & 0.981 $\pm$ 0.018 & 0.982 $\pm$ 0.017 & 0.870 $\pm$ 0.131 & 0.893 $\pm$ 0.117 \\
            Sparsity ($\downarrow$)    & 0.074 $\pm$ 0.035 & 0.033 $\pm$ 0.013 & 0.053 $\pm$ 0.009 & 0.402 $\pm$ 0.142 & 0.399 $\pm$ 0.151 \\
            Stability ($\downarrow$)   & 0.436 $\pm$ 0.165 & 0.300 $\pm$ 0.086 & 0.361 $\pm$ 0.092 & 0.931 $\pm$ 0.414 & 0.855 $\pm$ 0.396 \\
            \hline
        \end{tabular}
        \end{sc}
        \end{small}
    \end{center}
    \vskip -0.1in
\end{table*}

\begin{table*}[t!]
    \caption{Average interpretability metrics for models with three graph layers across different datasets. $\downarrow$ indicates that a lower metric value is better, while $\uparrow$ indicates that a higher value is better.}
    \label{tab:all_data_3l}
    \vskip 0.15in
    \begin{center}
        \begin{small}
        \begin{sc}
        \begin{tabular}{lccccc}
            \hline
            Dataset & Cora & CiteSeer & PubMed & Photo & Computers \\
            \hline
            Consistency ($\uparrow$) & 0.999 $\pm$ 0.001 & 0.997 $\pm$ 0.005 & 0.999 $\pm$ 0.003 & 1.000 $\pm$ 0.000 & 1.000 $\pm$ 0.001 \\
            Fidelity ($\uparrow$)    & 0.956 $\pm$ 0.023 & 0.961 $\pm$ 0.049 & 0.980 $\pm$ 0.017 & 0.983 $\pm$ 0.038 & 0.735 $\pm$ 0.173 \\
            Sparsity ($\downarrow$)  & 0.046 $\pm$ 0.005 & 0.043 $\pm$ 0.022 & 0.057 $\pm$ 0.009 & 0.231 $\pm$ 0.112 & 0.327 $\pm$ 0.136 \\
            Stability ($\downarrow$) & 0.366 $\pm$ 0.086 & 0.380 $\pm$ 0.144 & 0.347 $\pm$ 0.093 & 0.655 $\pm$ 0.290 & 1.271 $\pm$ 0.477 \\
            \hline
        \end{tabular}
        \end{sc}
        \end{small}
    \end{center}
    \vskip -0.1in
\end{table*}

As part of the first research question, the influence of domain characteristics and graph properties on interpretability metrics is analyzed to account for these factors in further evaluation. All models are grouped based on having two or three graph layers, respectively, and the results are averaged separately for each dataset. The results are presented in Tables \ref{tab:all_data_2l} and \ref{tab:all_data_3l}.

The conclusions drawn from the tables indicate that the metrics of \textbf{Consistency and Fidelity remained almost unchanged}, while the metrics of \textbf{Sparsity and Stability exhibited significant differences}. Datasets from the Amazon group have a significantly higher average degree compared to datasets from the Planetoid group, with fewer features. One more reason for this can be attributed to the domain characteristics, as the Amazon group datasets are partially constructed based on meta-information, which was originally represented in the form of images and text. This is a less structured form of representation compared to the original graph-based representation. In subsequent experiments, domains will be represented separately to ensure a more accurate comparison of methods and to avoid increasing dispersion during averaging.

It can also be noted that the metrics of Fidelity and Consistency either did not change or are comparable within the margin of error.

\subsubsection{Architecture's influence on interpretability}

Now, the impact of architectural decisions on model quality is compared. For this, data will be taken separately for each domain and averaged across all defense methods for each of the six architectures. The results are presented in Tables \ref{tab:all_models_pl} and \ref{tab:all_data_am}.

\begin{table*}[t!]
  \caption{Average interpretability metrics for different GNNs architectures across the Planetoid datasets group averaged across all defense methods. $\downarrow$ indicates that a lower metric value is better, while $\uparrow$ indicates that a higher value is better. Significant improvements in the corresponding metric are highlighted in bold, and significant degradations are italicised.}
  \label{tab:all_models_pl}
  \vskip 0.15in
  \begin{center}
    \begin{small}
      \begin{sc}
        \begin{tabular}{lcccccc}
          \hline
          Architecture & gat\_gat & gin\_gin & sage\_sage & gcn\_gcn & gcn\_gcn\_gcn & sage\_sage\_sage \\
          \hline
          consistency (\({\uparrow}\)) & 0.998 ± 0.001 & 0.997 ± 0.006 & 0.997 ± 0.004 & 0.999 ± 0.002 & 0.999 ± 0.001 & 0.998 ± 0.005 \\
          fidelity (\({\uparrow}\)) & 0.982 ± 0.017 & 0.974 ± 0.034 & 0.976 ± 0.020 & 0.980 ± 0.028 & 0.982 ± 0.017 & 0.950 ± 0.042 \\
          sparsity (\({\downarrow}\)) & 0.045 ± 0.007 & \textit{0.076 ± 0.018} & 0.046 ± 0.013 & 0.055 ± 0.014 & 0.045 ± 0.007 & \textbf{0.041 ± 0.012} \\
          stability (\({\downarrow}\)) & 0.370 ± 0.089 & \textit{0.494 ± 0.098} & \textbf{0.300 ± 0.079} & \textbf{0.296 ± 0.072} & \textbf{0.272 ± 0.048} & \textbf{0.257 ± 0.067} \\
          \hline
        \end{tabular}
      \end{sc}
    \end{small}
  \end{center}
  \vskip -0.1in
\end{table*}

\begin{table*}[t!]
  \caption{Average interpretability metrics for different GNNs architectures across the Amazon datasets group averaged across all defense methods. $\downarrow$ indicates that a lower metric value is better, while $\uparrow$ indicates that a higher value is better. Significant improvements in the corresponding metric are highlighted in bold, and significant degradations are italicised.}
  \label{tab:all_data_am}
  \vskip 0.15in
  \begin{center}
    \begin{small}
      \begin{sc}
        \begin{tabular}{lcccccc}
          \hline
          Architecture & gat\_gat & gin\_gin & sage\_sage & gcn\_gcn & gcn\_gcn\_gcn & sage\_sage\_sage \\
          \hline
          consistency (\({\uparrow}\)) & 0.998 ± 0.001 & 0.997 ± 0.004 & 0.998 ± 0.002 & 0.999 ± 0.001 & 1.000 ± 0.000 & 1.000 ± 0.001 \\
          fidelity (\({\uparrow}\)) & 0.899 ± 0.118 & 0.881 ± 0.117 & 0.849 ± 0.140 & 0.896 ± 0.122 & 0.885 ± 0.088 & 0.832 ± 0.124 \\
          sparsity (\({\downarrow}\)) & 0.388 ± 0.123 & \textit{0.525 ± 0.106} & 0.409 ± 0.103 & 0.394 ± 0.119 & \textbf{0.262 ± 0.09} & \textbf{0.317 ± 0.108} \\
          stability (\({\downarrow}\)) & 0.947 ± 0.262 & \textit{1.469 ± 0.470} & \textbf{0.481 ± 0.174} & \textbf{0.746 ± 0.25} & \textit{1.004 ± 0.296} & 0.921 ± 0.271 \\
          \hline
        \end{tabular}
      \end{sc}
    \end{small}
  \end{center}
  \vskip -0.1in
\end{table*}

The conclusions drawn from the tables indicate that the \textbf{GIN-based model} exhibits \textbf{significantly worse} values for the \textbf{Sparsity and Stability} metrics, while the \textbf{GCN and SAGE-based models} show \textbf{significantly better} values for these metrics. The deterioration in metrics for the GIN-based model can be explained by the method's focus on aligning the representations of isomorphic graphs, where the majority of a vertex's neighborhood is important. It is worth noting that removing even a single vertex can significantly reduce the isomorphism of the graphs, and the perturbed graph, when calculating the metric, will be located in a different region of the space, thus leading to a different interpretation. In contrast, GCN and SAGE convolutions operate on a simpler principle of aggregating information from neighbors, which explains their better stability metrics.

Additionally, it can be observed that increasing the number of layers has a positive effect on Sparsity but a negative effect on Stability. The first is logically explained by the fact that the neighborhood grows faster than the size of the interpretation in the larger neighborhood. On the other hand, small perturbations can have a stronger impact on more distant neighborhoods, potentially rendering them inaccessible due to random edge removal during graph perturbation. The larger the neighborhood, the greater the effect the perturbation has on it.

\subsubsection{Impact of defense mechanisms on interpretability}

In the final experiment, the impact of adding defense mechanisms against poisoning and evasion attacks on model interpretability is examined. Data will be taken separately for each domain and averaged across all architectures with the same number of layers. To reduce column width, defense methods will be denoted by the first letters of their names (JaccardDefense – JD, GNNGuard – GG, Gradient Regularization – GR, Defensive Distillation – DD, Adversarial Training – AT, Data Quantization Defense – DQD, Autoencoder Defense – AD). The results are presented in Tables \ref{tab:def-2l-pl}, \ref{tab:def-2l-am}, \ref{tab:def-3l-pl}, and \ref{tab:def-3l-am}.

\begin{table*}[t!]
  \caption{Average interpretability metrics for different defense mechanisms across the Planetoid datasets group with 2-layer architectures. Defense methods are denoted by their initials,  described in Section \ref{section_experiments}. $\downarrow$ indicates that a lower metric value is better, while $\uparrow$ indicates that a higher value is better. Significant improvements in the corresponding metric are highlighted in bold, and significant degradations are italicised.}
  \label{tab:def-2l-pl}
  \vskip 0.15in
  \begin{center}
    \begin{scriptsize}
      \begin{sc}
        \begin{tabular}{lcccccccc}
          \hline
          Metric & AT & AE & DD & GG & GR & JD & DQD & Unprotected \\
          \hline
          consistency (\({\uparrow}\)) & 1.000 ± 0.001 & 0.999 ± 0.002 & 0.999 ± 0.002 & 0.999 ± 0.003 & 1.000 ± 0.001 & 0.998 ± 0.006 & 0.999 ± 0.002 & 0.990 ± 0.007 \\
          fidelity (\({\uparrow}\)) & 0.998 ± 0.005 & 0.997 ± 0.009 & 0.999 ± 0.004 & 0.997 ± 0.009 & 0.998 ± 0.005 & 0.994 ± 0.012 & 0.997 ± 0.009 & 0.846 ± 0.143 \\
          sparsity (\({\downarrow}\)) & 0.010 ± 0.012 & 0.010 ± 0.012 & 0.007 ± 0.008 & 0.010 ± 0.012 & 0.010 ± 0.012 & 0.020 ± \textit{0.019} & 0.010 ± 0.012 & 0.347 ± 0.062 \\
          stability (\({\downarrow}\)) & 0.257 ± 0.097 & \textit{0.304 ± 0.077} & 0.177 ± 0.067 & 0.184 ± 0.081 & \textbf{0.160 ± 0.081} & \textbf{0.167 ± 0.057} & 0.201 ± 0.085 & 1.504 ± 0.313 \\
          \hline
        \end{tabular}
      \end{sc}
    \end{scriptsize}
  \end{center}
  \vskip -0.1in
\end{table*}

\begin{table*}[t!]
  \caption{Average interpretability metrics for different defense mechanisms across the Amazon datasets group with 2-layer architectures. Defense methods are denoted by their initials,  described in Section \ref{section_experiments}. $\downarrow$ indicates that a lower metric value is better, while $\uparrow$ indicates that a higher value is better. Significant improvements in the corresponding metric are highlighted in bold, and significant degradations are italicised.}
  \label{tab:def-2l-am}
  \vskip 0.15in
  \begin{center}
    \begin{scriptsize}
      \begin{sc}
        \begin{tabular}{lcccccccc}
          \hline
          Metric & AT & AE & DD & GG & GR & JD & DQD & Unprotected \\
          \hline
          consistency (\({\uparrow}\)) & 0.998 ± 0.003 & 0.999 ± 0.002 & 0.999 ± 0.002 & 0.999 ± 0.002 & 0.998 ± 0.004 & 0.998 ± 0.003 & 0.999 ± 0.001 & 0.997 ± 0.001 \\
          fidelity (\({\uparrow}\)) & 0.890 ± 0.119 & 0.905 ± 0.114 & 0.908 ± 0.113 & 0.887 ± 0.125 & 0.898 ± 0.112 & 0.885 ± 0.126 & 0.883 ± 0.127 & 0.796 ± 0.158 \\
          sparsity (\({\downarrow}\)) & 0.355 ± 0.156 & 0.342 ± 0.149 & 0.320 ± 0.151 & 0.343 ± 0.144 & 0.361 ± 0.138 & 0.306 ± \textit{0.196} & 0.359 ± 0.156 & 0.798 ± 0.074 \\
          stability (\({\downarrow}\)) & 0.622 ± 0.359 & \textit{1.014 ± 0.427} & 0.579 ± 0.298 & 0.714 ± 0.389 & 0.791 ± 0.403 & \textbf{0.458 ± 0.312} & 0.803 ± 0.437 & 2.227 ± 0.586 \\
          \hline
        \end{tabular}
      \end{sc}
    \end{scriptsize}
  \end{center}
  \vskip -0.1in
\end{table*}

\begin{table*}[t!]
  \caption{Average interpretability metrics for different defense mechanisms across the Planetoid datasets group with 3-layer architectures. Defense methods are denoted by their initials,  described in Section \ref{section_experiments}. $\downarrow$ indicates that a lower metric value is better, while $\uparrow$ indicates that a higher value is better. Significant improvements in the corresponding metric are highlighted in bold, and significant degradations are italicised.}
  \label{tab:def-3l-pl}
  \vskip 0.15in
  \begin{center}
    \begin{scriptsize}
      \begin{sc}
        \begin{tabular}{lcccccccc}
          \hline
          Metric & AT & AE & DD & GG & GR & JD & DQD & Unprotected \\
          \hline
          consistency (\({\uparrow}\)) & 0.999 ± 0.001 & 0.999 ± 0.002 & 0.999 ± 0.003 & 0.999 ± 0.003 & 0.999 ± 0.001 & 0.998 ± 0.005 & 0.999 ± 0.003 & 0.995 ± 0.004 \\
          fidelity (\({\uparrow}\)) & 0.995 ± 0.011 & 0.995 ± 0.011 & 0.995 ± 0.011 & 0.995 ± 0.011 & 0.995 ± 0.011 & 0.995 ± 0.011 & 0.995 ± 0.011 & 0.760 ± 0.160 \\
          sparsity (\({\downarrow}\)) & 0.007 ± 0.006 & 0.010 ± 0.007 & 0.006 ± 0.006 & 0.006 ± 0.006 & 0.005 ± 0.003 & 0.035 ± \textit{0.023} & 0.006 ± 0.006 & 0.350 ± 0.063 \\
          stability (\({\downarrow}\)) & 0.229 ± 0.048 & \textit{0.271 ± 0.073} & 0.170 ± 0.069 & 0.178 ± 0.068 & 0.150 ± 0.065 & \textbf{0.120 ± 0.095} & 0.180 ± 0.074 & 1.607 ± 0.312 \\
          \hline
        \end{tabular}
      \end{sc}
    \end{scriptsize}
  \end{center}
  \vskip -0.1in
\end{table*}

\begin{table*}[t!]
  \caption{Average interpretability metrics for different defense mechanisms across the Amazon datasets group with 3-layer architectures. Defense methods are denoted by their initials,  described in Section \ref{section_experiments}. $\downarrow$ indicates that a lower metric value is better, while $\uparrow$ indicates that a higher value is better. Significant improvements in the corresponding metric are highlighted in bold, and significant degradations are italicised.}
  \label{tab:def-3l-am}
  \vskip 0.15in
  \begin{center}
    \begin{scriptsize}
      \begin{sc}
        \begin{tabular}{lcccccccc}
          \hline
          Metric & AT & AE & DD & GG & GR & JD & DQD & Unprotected \\
          \hline
          consistency (\({\uparrow}\)) & 1.000 ± 0.001 & 1.000 ± 0.000 & 1.000 ± 0.001 & 1.000 ± 0.000 & 1.000 ± 0.000 & 1.000 ± 0.001 & 1.000 ± 0.000 & 0.999 ± 0.000 \\
          fidelity (\({\uparrow}\)) & 0.843 ± 0.108 & 0.849 ± 0.106 & 0.849 ± 0.106 & 0.843 ± 0.108 & 0.856 ± 0.105 & 0.850 ± 0.106 & 0.856 ± 0.105 & 0.925 ± 0.101 \\
          sparsity (\({\downarrow}\)) & 0.241 ± 0.139 & 0.235 ± 0.112 & 0.235 ± 0.121 & 0.241 ± 0.129 & 0.235 ± 0.125 & 0.241 ± \textit{0.182} & 0.235 ± 0.136 & 0.574 ± 0.099 \\
          stability (\({\downarrow}\)) & 0.713 ± 0.349 & \textit{0.991 ± 0.383} & 0.738 ± 0.360 & 0.753 ± 0.368 & 0.781 ± 0.405 & \textbf{0.482 ± 0.384} & 0.763 ± 0.386 & 2.436 ± 0.430 \\
          \hline
        \end{tabular}
      \end{sc}
    \end{scriptsize}
  \end{center}
  \vskip -0.1in
\end{table*}

The conclusions drawn from the tables indicate that \textbf{all examined defense mechanisms improve interpretability metrics} compared to the unprotected model. The most likely explanation is that these defense mechanisms operate on mathematical principles similar to regularization, which often enhances the final model.

One notable exception is \textbf{adversarial training}, which \textbf{badly impacts Stability}. A possible reason is that generating adversarial examples and training on them results in a highly complex decision boundary between classes, making the interpretation method less stable when small perturbations are introduced.

Another noteworthy observation is the effect of the \textbf{Jaccard defense} method, which \textbf{improves Stability but significantly increases the variance of the Sparsity} metric. This method removes suspicious and low-quality edges, which benefits the final model under small deviations. However, since it eliminates some edges, its impact on Sparsity is not straightforward. In some cases, a large number of edges may be removed from a node’s neighborhood, making the number of important edges nearly equal to the total number of edges in the neighborhood, leading to an increase in the metric. In other cases, the neighborhood may shrink only slightly, but the interpretation method selects fewer important edges, assigning them higher importance. These opposing effects cause the variance of the Sparsity metric to increase. The Appendix \ref{app:subx} presents the results of the interpretation metrics using the method SubgraphX.

\section{Discussion}
\label{section_discussion}

Based on all experiments, more general conclusions can be drawn. The Consistency and Fidelity metrics appear to be better suited for evaluating interpretation methods, as they are less affected by factors such as the addition of defense mechanisms and architectural decisions. In contrast, the Sparsity and Stability metrics are more appropriate for analyzing how small changes impact model interpretability.

A key result is that the addition of most popular defense methods improves model interpretability. However, despite these findings based on the considered methods, it is important to emphasize that interpretability evaluation metrics should be further refined—primarily to establish a clear understanding of when and how each metric should be applied. Currently, these metrics either remain almost unaffected by model modifications or, conversely, are influenced by too many factors, making it difficult to determine how a specific modification to the model or its usage pipeline impacts overall interpretability.

\section{Conclusion}
\label{section_conclusion}

This paper introduces a comprehensive benchmark for analyzing the impact of various factors on the interpretability of GNNs, particularly under adversarial conditions such as poisoning and evasion attacks. We highlight the complex relationship between robustness and interpretability evaluating multiple GNN architectures and defense mechanisms across different datasets. Our findings show that while most defense mechanisms enhance interpretability, their effects vary depending on the architecture and the specific defense applied. Additionally, we identify the limitations of current interpretability metrics, suggesting that refinements are necessary for capturing the nuanced impact of model modifications. This benchmark provides a foundation for developing GNNs robust to adversarial threats and interpretable, promoting their use in sensitive and high-stakes domains. The framework will be made publicly available, offering a valuable tool for future research in this area.

% \section*{Acknowledgements}

% //

\section*{Impact Statement}
This paper presents work whose goal is to advance the field
of Machine Learning. There are many potential societal
consequences of our work, none which we feel must be
specifically highlighted here

\nocite{langley00}

\bibliography{reference}
\bibliographystyle{icml2025}

\newpage
\appendix
\onecolumn

\section{Datasets statistic}
\label{app:datasets}

Statistics of the data sets used in the experiments are presented in Table \ref{tab:dataset_statistics}.  Cora, CiteSeer, and PubMed, which belong to the citation domain \cite{sen2008collective} and are included in the Torch-Geometric library under the Planetoid dataset group, as well as Computers and Photo from the purchase graph domain \cite{mcauley2015image}, which are also available in PyTorch-Geometric under the Amazon dataset group. 

\section{GNNs architectures}
\label{app:models}

\begin{table}[t]
    \caption{Statistics of the datasets used in the experiments.}
    \label{tab:dataset_statistics}
    \vskip 0.15in
    \begin{center}
        \begin{small}
        \begin{sc}
        \begin{tabular}{lcccc}
            \hline
            Dataset   & Nodes  & Edges   & Classes & Features \\
            \hline
            Cora      & 2,708  & 10,556  & 7       & 1,433    \\
            CiteSeer  & 3,327  & 9,104   & 6       & 3,703    \\
            PubMed    & 19,717 & 88,648  & 3       & 500      \\
            Computers & 13,752 & 491,722 & 10      & 767      \\
            Photo     & 7,650  & 238,162 & 7       & 745      \\
            \hline
        \end{tabular}
        \end{sc}
        \end{small}
    \end{center}
    \vskip -0.1in
\end{table}
 
\begin{verbatim}
GCN-2l(
  Sequential(
    (0): GCNConv(input_size, 16)
    (1): ReLU(inplace)
    (2): GCNConv(16, output_size)
    (3): LogSoftmax(inplace)
  )
)
\end{verbatim}

\begin{verbatim}
GCN-3l(
  Sequential(
    (0): GCNConv(input_size, 16)
    (1): ReLU(inplace)
    (2): GCNConv(16, 16)
    (3): ReLU(inplace)
    (4): GCNConv(16, output_size)
    (5): LogSoftmax(inplace)
  )
)
\end{verbatim}

\begin{verbatim}
SAGE-2l(
  Sequential(
    (0): SAGEConv(input_size, 16)
    (1): BatchNorm1d(16, 16, eps=1e-05)
    (2): ReLU(inplace)
    (3): SAGEConv(16, output_size)
    (4): LogSoftmax(inplace)
  )
)
\end{verbatim}

\begin{verbatim}
SAGE-3l(
  Sequential(
    (0): SAGEConv(input_size, 16)
    (1): BatchNorm1d(16, 16, eps=1e-05)
    (2): ReLU(inplace)
    (3): SAGEConv(16, 16)
    (4): BatchNorm1d(16, 16, eps=1e-05)
    (5): ReLU(inplace)
    (6): SAGEConv(16, output_size)
    (7): LogSoftmax(inplace)
  )
)
\end{verbatim}

\begin{verbatim}
GIN-2l( 
    Sequential( 
        (0): GINConv( 
            Sequential( 
                (0): Linear(input_size, 16) 
                (1): BatchNorm1d(16, eps=1e-05) 
                (2): ReLU(inplace) 
                (3): Linear(16, 16) 
                (4): BatchNorm1d(16, eps=1e-05) 
                (5): ReLU(inplace) 
            ) 
        ) 
        (1): ReLU(inplace) 
        (2): GINConv( 
            Sequential( 
                (0): Linear(16, 16) 
                (1): BatchNorm1d(16, eps=1e-05) 
                (2): ReLU(inplace) 
                (3): Linear(16, output_size) 
            ) 
        ) 
        (3): LogSoftmax(inplace) 
    ) 
)
\end{verbatim}

\begin{verbatim}
GAT-2l(
  Sequential(
    (0): GATConv(input_size, 16, heads=3)
    (1): BatchNorm1d(48, 48, eps=1e-05)
    (2): ReLU(inplace)
    (3): GATConv(48, output_size, heads=1)
    (4): LogSoftmax(inplace)
  )
)
\end{verbatim}

\section{Interpretation and defense methods formalization}
\label{app:int-def}

This appendix provides a more detailed description of all the interpretation and defense methods used in the experiments. The hyperparameters for the interpretation and defense methods are presented in  Table \ref{tab:hyper-int-def-full}

\subsection{Interpretation methods}
\subsubsection{GNNExpainer}
The GNNExplainer~\cite{ying2019gnnexplainer} algorithm aims to find a subgraph $G_S$ within a computational graph $G$ that maximizes the mutual information between two random variables --- specifically, the difference in entropy between the model's prediction and the conditional entropy given the subgraph. This optimization problem can be formulated as:
$$
\max_{G_S} MI(Y,G_S) = H(Y) - H(Y|G = G_S)
$$
Since $H(Y)$ is constant, maximizing this expression requires minimizing the conditional entropy $H(Y|G = G_S)$.
The conditional entropy $H(Y|G = G_S)$ can be expressed as the conditional expectation of the log probability of the prediction given the subgraph $G_S$. However, solving this problem by directly enumerating all possible subgraphs is computationally inefficient due to the exponentially large number of subgraphs. Therefore, a differentiable mask $M$ over the edges of the subgraph is trained using gradient descent. The optimization task is then written as:
$$
\min_{M} - \sum_{i=1}^C \mathbbm{1}[y = i] \log P_{\Phi}(y|A_G \odot \sigma(M))
$$
where $A_G$ is the adjacency matrix of the computational graph, and $\sigma(M)$ applies a sigmoid transformation to the mask. After training the mask, low-value elements are removed to obtain the final explanation for the model’s prediction.

\subsubsection{SubgrapghX}
SubgraphX~\cite{yuan2021explainability} generates a connected subgraph $\mathcal{G}^*$ of the computational graph $\mathcal{G}$. Let $\{\mathcal{G}_1,...,\mathcal{G}_i,...,\mathcal{G}_n\}$ represent all possible connected subgraphs of the computational graph, and let $f(\cdot)$ denote the model. The important subgraph $\mathcal{G}^*$ is then selected as: $\mathcal{G}^*$ = argmax Score$(f(\cdot),\mathcal{G},\mathcal{G}_i)$, where the Score function uses the Shapley value $\phi(\mathcal{G}_i)$ defined as:
$$
\phi(\mathcal{G}_i) = \sum_{S \subseteq P^{'} \setminus \{\mathcal{G}_i\}} \frac{|S|!(|P^{'}|-|S|-1)!}{|P^{'}|!} m(S, \mathcal{G}_i)
$$
Here, $m(S, \mathcal{G}_i) = f(S \cup \{\mathcal{G}_i\}) - f(S)$, $S$ is the coalition value, and $P^{'}$ denotes the set of vertices in the computational subgraph. In the case of large computational subgraphs, SubgraphX uses Monte Carlo tree search to explore the subgraph effectively.
\subsection{Defense methods}
\subsubsection{JaccardDefense}
JaccardDefense~\cite{wu2019adversarial} is a poison defender that removes edges between nodes that are not similar concerning the Jaccard Index (binary features of nodes being compared). This is followed by the idea that many attack methods are trying to connect not similar nodes to shadow important links.

\subsubsection{GNNGuard}
GNNGuard~\cite{zhang2020gnnguard} is a poison defender that diminishes message flow from suspicious edges by additional defense coefficients. With the use of the message-passing paradigm, GNN can be represented as: $$f = ({MSG}, {AGG}, {UPD}),$$ where ${MSG}$ - Message-passing function that specifies information message $m_{uv}^k$ transferred from node $u$ to $v$, ${AGG}$ - aggregates messages over node neighborhood, ${UPD}$ - combines aggregated message and embedding for layer $k$ to derive embedding of $k+1$ layer: $h_u^{k+1} = {UPD}(h_u^k, \hat{m}_u^k)$
\par According to this representation, GNNGuard modifies ${AGG}$ and ${UPD}$: every aggregated message $m_{uv}^k$ being transformed: ${m_{uv}^k}^{'} = m_{uv}^k \odot w_{uv}^k$ and within modified ${UPD}$ transformed ${h_u^k}^{'} = h_u^k \odot w_{uv}^k$ being combined with ${m_{uv}^k}^{'}$ ($\odot$ denotes dot product). These defense weights $w_{uv}^k$ are jointly learned with network parameters.

\subsubsection{Gradient Regularization}
Also Modifying the loss function used by a model with additional gradient regularization~\cite{finlay2021scaleable} can serve as a defense method.
$$
L = l(f(x),y) + \lambda (\frac{1}{h^2n}\norm{f(z) - f(x)}_2^2),
$$
where
$$
z = x + h \ \frac{\nabla l(f(x),y)}{\norm{\nabla l(f(x),y)}_2},
$$
$h$ is a quantization step and $\lambda$ is a regularization coefficient.

\subsubsection{Defensive Distillation}
Distillation as a defense method is about creating a copy of the original model that is more robust to attacks. The new model uses the original one as a teacher and so-called smooth labels for this model are obtained using the $softmax(x, T)$ on the last layer of the teacher:
$$
{{softmax}(x,T)}_i = \frac{e^{x_i/T}}{\sum_j {e^{x_j/T}}}
$$

\subsubsection{Adversarial Training}
Adversarial training~\cite{goodfellow2014explaining} is a defense technique that implies adding adversarial examples in a train set. Therefore an adversarial part is added to the loss function:
$$
L = l(f(x),y) + \lambda \ l(f(x^{'},y)),
$$
where $x^{'}$ is adversarial sample and $\lambda$ is adversarial training coefficient.

\subsubsection{Data Quantization Defense}
Quantization is a preprocessing technique that transforms continuous values into discrete values arranged on a uniform grid~\cite{guo2017countering}. While this method may diminish the quality of the original data, it can effectively mitigate the effects of adversarial attacks. Consequently, the fault diagnosis model needs to be retrained using the quantized data.

\subsubsection{Autoencoder Defense}
Autoencoders can be used to perform robust training too as it was shown in~\cite{meng2017magnet}. In this case, minimized loss function:
$$
L = \norm{x_{AE} - x}_1,
$$
where $x_{AE} = {autoencoder}(x+\epsilon)$ is a reconstructed data and $\epsilon$ is added noise.

\begin{table}[t]
    \caption{The hyperparameters for the interpretation and defense methods.}
    \label{tab:hyper-int-def-full}
    \vskip 0.15in
    \begin{center}
        \begin{small}
        \begin{sc}
        \begin{tabular}{@{}ll@{}}
            \toprule
            \textbf{Method} & \textbf{Hyperparameters} \\ \midrule
            GNNExplainer & 
            \begin{tabular}[c]{@{}l@{}}
                epochs = $100$ \\
                lr = $0.01$ \\
                node\_mask\_type = \texttt{attributes} \\
                edge\_mask\_type = \texttt{None} \\
                mode = \texttt{multiclass\_classification} \\
                return\_type = \texttt{log\_probs} \\
                edge\_size = $0.005$ \\
                edge\_reduction = \texttt{sum} \\
                node\_feat\_size = $1$ \\
                node\_feat\_reduction = \texttt{mean} \\
                edge\_ent = $1$ \\
                node\_feat\_ent = $0.1$ \\
                EPS = $1 \times 10^{-15}$ 
            \end{tabular} \\ \midrule
            
            SubgraphX & 
            \begin{tabular}[c]{@{}l@{}}
                rollout = $20$ \\
                min\_atoms = $5$ \\
                c\_puct = $10$ \\
                expand\_atoms = $14$ \\
                local\_radius = $4$ \\
                sample\_num = $100$ \\
                reward\_method = \texttt{mc\_l\_shapley} \\
                high2low = false \\
                subgraph\_building\_method = \texttt{zero\_filling} \\
                max\_nodes = $5$
            \end{tabular} \\ \midrule
            
            Jaccard & 
            \begin{tabular}[c]{@{}l@{}}
                threshold = $0.4$ 
            \end{tabular} \\ \midrule

            GNNGuard & 
            \begin{tabular}[c]{@{}l@{}}
                lr = $0.01$ \\
                attention = true \\
                drop = true \\
                train\_iters = $50$ 
            \end{tabular} \\ \midrule
            
            Adversarial training & 
            \begin{tabular}[c]{@{}l@{}}
                attack\_name = $FGSM$ \\
                $\epsilon = 0.01$ 
            \end{tabular} \\ \midrule
            
            Gradient Regularization & 
            \begin{tabular}[c]{@{}l@{}}
                regularization\_strength = $50$ 
            \end{tabular} \\ \midrule

            Distillation Defender & 
            \begin{tabular}[c]{@{}l@{}}
                temperature = $5$ 
            \end{tabular} \\ \midrule

            Quantization Defender & 
            \begin{tabular}[c]{@{}l@{}}
                num\_levels = $8$ 
            \end{tabular} \\ \midrule

            Autoencoder Defender & 
            \begin{tabular}[c]{@{}l@{}}
                hidden\_dim = $7$ \\
                bottleneck\_dim = $5$ \\
                reconstruction\_loss\_weight = $0.1$
            \end{tabular} \\ 
            
            \bottomrule
        \end{tabular}
        \end{sc}
        \end{small}
    \end{center}
    \vskip -0.1in
\end{table}

\section{ Results of experiments with SubgraphX}
\label{app:subx}

This appendix will describe short experiments with the SubgraphX method.

Since the computation time even on the Cora and simple architectures sometimes reached several hours on one vertex, the result was averaged based on 5 iterations and 5 vertices within each iteration. Dataset only Cora, models GCN-2l and GCN-3l.

The results are presented in Tables \ref{tab:subx-data}, \ref{tab:subx-models} and \ref{tab:subx-def}.

From the tables it is clear that the metrics are smaller on average, but all the key conclusions are also valid when using another post-hoc interpretation method.

\begin{table*}[t!]
  \caption{Average interpretability metrics for GCN-based models on the Cora dataset use SubgraphX. $\downarrow$ indicates that a lower metric value is better, while $\uparrow$ indicates that a higher value is better.}
  \label{tab:subx-data}
  \vskip 0.15in
  \begin{center}
    \begin{small}
      \begin{sc}
        \begin{tabular}{lc}
            \hline
            Dataset & Cora \\
            \hline
            Consistency ($\uparrow$) & 0.937 $\pm$ 0.040 \\
            Fidelity ($\uparrow$)    & 1.000 $\pm$ 0.000 \\
            Sparsity ($\downarrow$)  & 0.506 $\pm$ 0.138 \\
            Stability ($\downarrow$) & 2.205 $\pm$ 0.716 \\
            \hline
        \end{tabular}
      \end{sc}
    \end{small}
  \end{center}
  \vskip -0.1in
\end{table*}

\begin{table*}[t!]
  \caption{Average interpretability metrics for different GNN architectures on the Cora dataset use SubgraphX. $\downarrow$ indicates that a lower metric value is better, while $\uparrow$ indicates that a higher value is better.}
  \label{tab:subx-models}
  \vskip 0.15in
  \begin{center}
    \begin{small}
      \begin{sc}
        \begin{tabular}{lcc}
            \hline
            Architecture & gcn\_gcn & gcn\_gcn\_gcn \\
            \hline
            Consistency ($\uparrow$) & 0.966 $\pm$ 0.032 & 0.707 $\pm$ 0.101 \\
            Fidelity ($\uparrow$)    & 1.000 $\pm$ 0.000 & 1.000 $\pm$ 0.000 \\
            Sparsity ($\downarrow$)  & 0.462 $\pm$ 0.147 & 0.862 $\pm$ 0.061 \\
            Stability ($\downarrow$) & 2.047 $\pm$ 0.725 & 3.467 $\pm$ 0.642 \\
            \hline
        \end{tabular}
      \end{sc}
    \end{small}
  \end{center}
  \vskip -0.1in
\end{table*}

\begin{table*}[t!]
  \caption{Average interpretability metrics for different defense mechanisms for GNN across the Cora datasets use SubgraphX. Defense methods are denoted by their initials,  described in Section \ref{section_experiments}. $\downarrow$ indicates that a lower metric value is better, while $\uparrow$ indicates that a higher value is better.}
  \label{tab:subx-def}
  \vskip 0.15in
  \begin{center}
    \begin{scriptsize}
      \begin{sc}
        \begin{tabular}{lcccccccc}
            \hline
            Defense Method & AT & AE & DD & GG & GR & JD & DQD & Unprotected \\
            \hline
            Consistency ($\uparrow$) & 0.996 $\pm$ 0.012 & 0.985 $\pm$ 0.024 & 0.974 $\pm$ 0.029 & 0.981 $\pm$ 0.028 & 0.984 $\pm$ 0.021 & 0.924 $\pm$ 0.062 & 0.990 $\pm$ 0.019 & 0.799 $\pm$ 0.081 \\
            Fidelity ($\uparrow$)    & 1.000 $\pm$ 0.000 & 1.000 $\pm$ 0.000 & 1.000 $\pm$ 0.000 & 1.000 $\pm$ 0.000 & 1.000 $\pm$ 0.000 & 1.000 $\pm$ 0.000 & 1.000 $\pm$ 0.000 & 1.000 $\pm$ 0.000 \\
            Sparsity ($\downarrow$)  & 0.418 $\pm$ 0.139 & 0.405 $\pm$ 0.164 & 0.382 $\pm$ 0.158 & 0.414 $\pm$ 0.140 & 0.432 $\pm$ 0.142 & 0.519 $\pm$ \textit{0.197} & 0.402 $\pm$ 0.152 & 0.801 $\pm$ 0.084 \\
            Stability ($\downarrow$) & 2.068 $\pm$ 0.744 & \textit{2.450 $\pm$ 0.726} & 2.190 $\pm$ 0.640 & 1.885 $\pm$ 1.047 & 1.660 $\pm$ 0.676 & \textbf{1.553 $\pm$ 0.572} & 1.958 $\pm$ 0.730 & 3.340 $\pm$ 0.649 \\
            \hline
        \end{tabular}
      \end{sc}
    \end{scriptsize}
  \end{center}
  \vskip -0.1in
\end{table*}

\end{document}